\documentclass[10pt,twocolumn,letterpaper]{article}

\usepackage{iccv}              

\usepackage[accsupp]{axessibility}  

\definecolor{iccvblue}{rgb}{0.21,0.49,0.74}
\usepackage[pagebackref,breaklinks,colorlinks,allcolors=iccvblue]{hyperref}
\usepackage{multirow}
\def\paperID{14957} 
\def\confName{ICCV}
\def\confYear{2025}

\title{ForeSight: Multi-View Streaming \\Joint Object Detection and Trajectory Forecasting}

\author{Sandro Papais \qquad Letian Wang \qquad Brian Cheong \qquad Steven L. Waslander\\
University of Toronto \\
{\tt\small \{sandro.papais, letian.wang, brian.cheong, steven.waslander\}@robotics.utias.utoronto.ca}
}

\begin{document}
\maketitle
\begin{abstract}
We introduce ForeSight, a novel joint detection and forecasting framework for vision-based 3D perception in autonomous vehicles. Traditional approaches treat detection and forecasting as separate sequential tasks, limiting their ability to leverage temporal cues. ForeSight addresses this limitation with a multi-task streaming and bidirectional learning approach, allowing detection and forecasting to share query memory and propagate information seamlessly. The forecast-aware detection transformer enhances spatial reasoning by integrating trajectory predictions from a multiple hypothesis forecast memory queue, while the streaming forecast transformer improves temporal consistency using past forecasts and refined detections. Unlike tracking-based methods, ForeSight eliminates the need for explicit object association, reducing error propagation with a tracking-free model that efficiently scales across multi-frame sequences. Experiments on the nuScenes dataset show that ForeSight achieves state-of-the-art performance, achieving an EPA of 54.9\%, surpassing previous methods by 9.3\%, while also attaining the best mAP  and minADE among multi-view detection and forecasting models.\footnote{Project page: \url{https://foresight-iccv.github.io}}
\end{abstract}    
\section{Introduction}
\label{sec:intro}

Perception systems in autonomous vehicles (AVs) are critical for understanding the dynamic driving environment. Vision-based 3D detection has gained popularity due to its low deployment cost and ability to interpret visual elements. Despite this progress, vision-based systems struggle in complex driving scenarios, particularly with occluded or partially visible objects which can pose critical safety risks. To address these challenges, temporal learning methods that leverage multi-frame histories have emerged as a solution, improving detection by aggregating past observations and motion cues.

\begin{figure*}[!t]
  \centering
  \includegraphics[width=\textwidth]{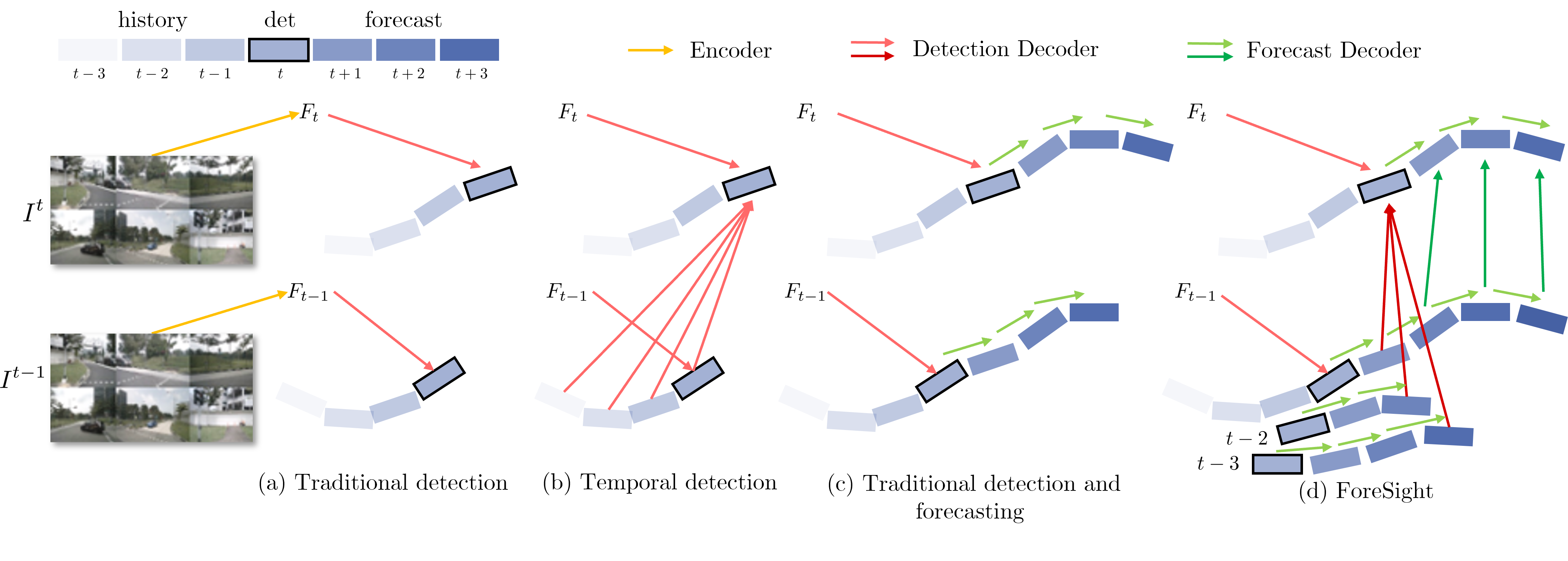}
  \vspace{-0.1in}
  \caption{A comparison of temporal learning perception methods related to ForeSight. (a) Traditional object detection methods leverage single-frame sensor data. (b) Temporal sparse detectors typically keep memory of past detection locations without explicit forecasting capabilities. (c) Typical joint detection-forecasting works pass information from detection to forecasting and rely on tracking for temporal information. (d) ForeSight further propagates motion forecast information to future frames to be reused in detection and forecasting.}
  \label{fig:overview}
\end{figure*}


Existing temporal learning methods fall into two categories: \textit{dense-feature} and \textit{sparse-query} learning. Dense-feature methods~\cite{li2022bevformer, park2022time, li2023bevdepth, huang2022bevdet4d, kim2023predict} align historic bird's-eye-view (BEV) features to the current frame for fusion. Sparse-query methods~\cite{liu2022petr, liu2023petrv2, wang2023exploring, lin2023sparse4d} represent objects as sparse queries and use attention~\cite{vaswani2017attention} to selectively fuse temporal features, achieving strong efficiency and performance. However, they often overlook the multi-modal motion of surrounding agents, limiting their ability to fully exploit temporal information. Given advances in motion forecasting~\cite{zhou2023query, varadarajan2022multipath++, zhou2024smartrefine, wang2022transferable}, neglecting to fully incorporate future movement in detection discards valuable cues.

Traditional motion forecasting models rely on curated ground-truth trajectories, assuming perfect detection and tracking. However, this assumption does not hold in real-world, online perception scenarios, leading to degraded performance~\cite{xu2024towards}. Meanwhile, emerging end-to-end methods integrating perception and forecasting architectures into joint training show promise in the presence of perception errors \cite{hu2023planning,gu2023vip3d}. These methods rely on tracking to bridge the gap between forecasting and perception. However, they trade off detection accuracy for tracking performance and introduce errors from incorrect trajectories association, bottlenecking learning. Moreover, these forecasting methods recompute trajectories at each timestep, making them computationally inefficient. By integrating forecasting with temporal detection through a streaming memory queue, ForeSight enhances computational efficiency and ensures flexible propagation of historical context.

Our key insight is to re-think the traditional autonomy stack, where motion forecasting is typically used for planning and then discarded. Instead, we propose an integrated system where motion forecasts are fed back into both detection and forecasting via a joint memory queue. Therefore, we introduce \textbf{ForeSight}, a novel framework that jointly optimizes detection and forecasting through unified process for query initialization, attention, and propagation. Experiments show that bidirectional query flow enhances both tasks, achieving state-of-the-art end-to-end forecasting performance. ForeSight uses a unified transformer architecture for temporal detection and track-free forecasting with a joint memory system, allowing past detections and forecasts to serve as a prior on the current detections and forecasts. To our knowledge, ForeSight is the first joint perception and forecasting method to propagate all queries bidirectionally with closed-loop feedback. ForeSight represents a promising advance toward improved memory systems for efficient architectures for autonomous systems. To summarize, our key contributions are: 
\begin{itemize}
    \item We introduce ForeSight, a joint streaming detection and forecasting framework for multi-view images. Unlike existing methods that treat these as sequential tasks, ForeSight enables both to share temporal priors via a unified memory, enhancing performance of both tasks. Our bidirectional query propagation allows information to flow not just from detection to forecasting, as in conventional autonomy stacks, but also back from forecasting to both tasks, enriching detection with temporal cues.
    \item We introduce a tracking-free streaming forecast approach to improve direct integration between detection and forecasting by removing the tracking bottleneck and adding a shared memory access. We find the forecasting model benefits from this approach, leading to improved end-to-end forecasting.
    \item Extensive experiments on the NuScenes dataset \cite{caesar2020nuscenes} demonstrate that ForeSight outperforms existing perception and forecasting methods, achieving the highest EPA by a large margin of 9.3\% over SOTA joint perception and forecasting method UniAD\cite{hu2023planning}. ForeSight also surpasses the best multi-view 3D detection baseline, StreamPETR, with a 2.1\% mAP improvement.
\end{itemize}

\section{Related Works}
\label{sec:related_works}

\subsection{Multi-View 3D Object Detection}
Multi-view 3D object detection identifies objects in 3D space by transforming 2D image features from multiple viewpoints into a unified 3D representation. BEV-based methods project 2D features onto a BEV plane via depth estimation, as in LSS~\cite{philion2020lift}, CaDDN~\cite{reading2021categorical}, and BEVDet \cite{huang2021bevdet}. Sparse query-based methods, including DETR3D \cite{wang2022detr3d} and PETR \cite{liu2022petr}, use learnable 3D queries to aggregate multi-view features via attention~\cite{vaswani2017attention}. ForeSight follows the sparse query-based approach for object detection.

\subsection{Temporal Multi-View 3D Object Detection}
Temporal fusion improves single-frame detection by aggregating features across time. Dense BEV-based methods like BEVDet4D~\cite{huang2022bevdet4d} and Fiery~\cite{hu2021fiery} warp and concatenate dense BEV features from past frames, improving detection stability and accuracy. SOLOFusion~\cite{park2022time} added caching features across a longer sliding window. BEVFormer~\cite{li2022bevformer} improved feature aggregation with deformable attention. 

Sparse query-based methods, such as StreamPETR~\cite{wang2023exploring}, PETRv2~\cite{liu2023petrv2}, and Sparse4Dv2~\cite{lin2023sparse4d}, cache object queries from past frames to efficiently capture temporal dependencies. These approaches attend to past detection and initialize new queries from them to reduce computational overhead compared to dense BEV-based methods. However, they lack explicit supervision for propagating detections into the future, limiting temporal learning. ForeSight addresses this by jointly learning detection and forecasting, explicitly supervising the influence of both histories.

\subsection{Motion Forecasting}
Motion forecasting for AVs predicts future agent positions based on the road map, agent history, and semantic class. Early works~\cite{cui2019multimodal,chai2019multipath,salzmann2020trajectron++} rasterize scene context into bird-eye-view images for CNN processing. Later methods adopt vector-based encodings with permutation-invariant operators such as pooling~\cite{varadarajan2022multipath++}, graph convolutions~\cite{gao2020vectornet,liang2020learning}, and attention mechanisms~\cite{wang2023prophnet,li2024scenarionet,feng2023trafficgen,zhou2024smartrefine}. Alternative scene representations, including dynamic insertion areas~\cite{wang2021hierarchical} and Frenet frame-based approaches~\cite{ye2023improving,wang2021socially,wang2023efficient}, improve generalizability. However, most assume noise-free inputs, leading to degradation in real-world settings. Some works address uncertainties in object classes~\cite{ivanovic2022heterogeneous}, map elements~\cite{gu2024accelerating}, and object tracks~\cite{weng2022whose,weng2022mtp}, but focus solely on forecasting task trained from curated ground truth.

\subsection{Joint Detection and Forecasting}
Training motion forecasting separately from detection on curated data degrades real-world performance~\cite{xu2024towards}. Joint training mitigates this issue by improving the information flow between both tasks. Early works such as FaF \cite{luo2018fast}, IntentNet \cite{casas2018intentnet}, and PnPNet \cite{liang2020pnpnet} demonstrate the benefits over standalone conventional models. Recent methods, such as InterFuser \cite{shao2023safety}, VAD \cite{jiang2023vad}, ReasonNet \cite{shao2023reasonnet} and LMDrive \cite{shao2024lmdrive} have extended this idea to add downstream tasks such as planning and control. Some incorporate tracking for temporal cues, either jointly~\cite{pang2023standing} or sequentially~\cite{papais2024swtrack}. End-to-end joint training of the full set of driving tasks has gained traction, with UniAD~\cite{hu2023planning} and ViP3D~\cite{gu2023vip3d} integrating sparse queries across all tasks. However, most of these recent methods enforce tracking via ground truth query assignments which introduces information bottlenecks \cite{casas2024detra} and results in poorer detection performance \cite{cheong2024jdt3d}.

While end-to-end methods only consider flow of information from detection to forecasting, recent works explore using forecasting for detection. FaF \cite{luo2018fast} and FrameFusion \cite{li2024frame} showed that past detection boxes forecast to present and averaged with current detections improves detection performance. P2D~\cite{kim2023predict} explored prediction for feature aggregation and MoDAR \cite{li2023modar} explored forecasting previous sensor data as additional input to the detection model. However, these methods are not jointly trained to perform detection and forecasting. ForeSight introduces a streaming joint detection-forecasting framework with bidirectional propagation, enabling closed-loop training for both tasks.
\section{Method}
\label{sec:method}

\begin{figure*}[!ht]
  \centering
  \includegraphics[width=\textwidth]{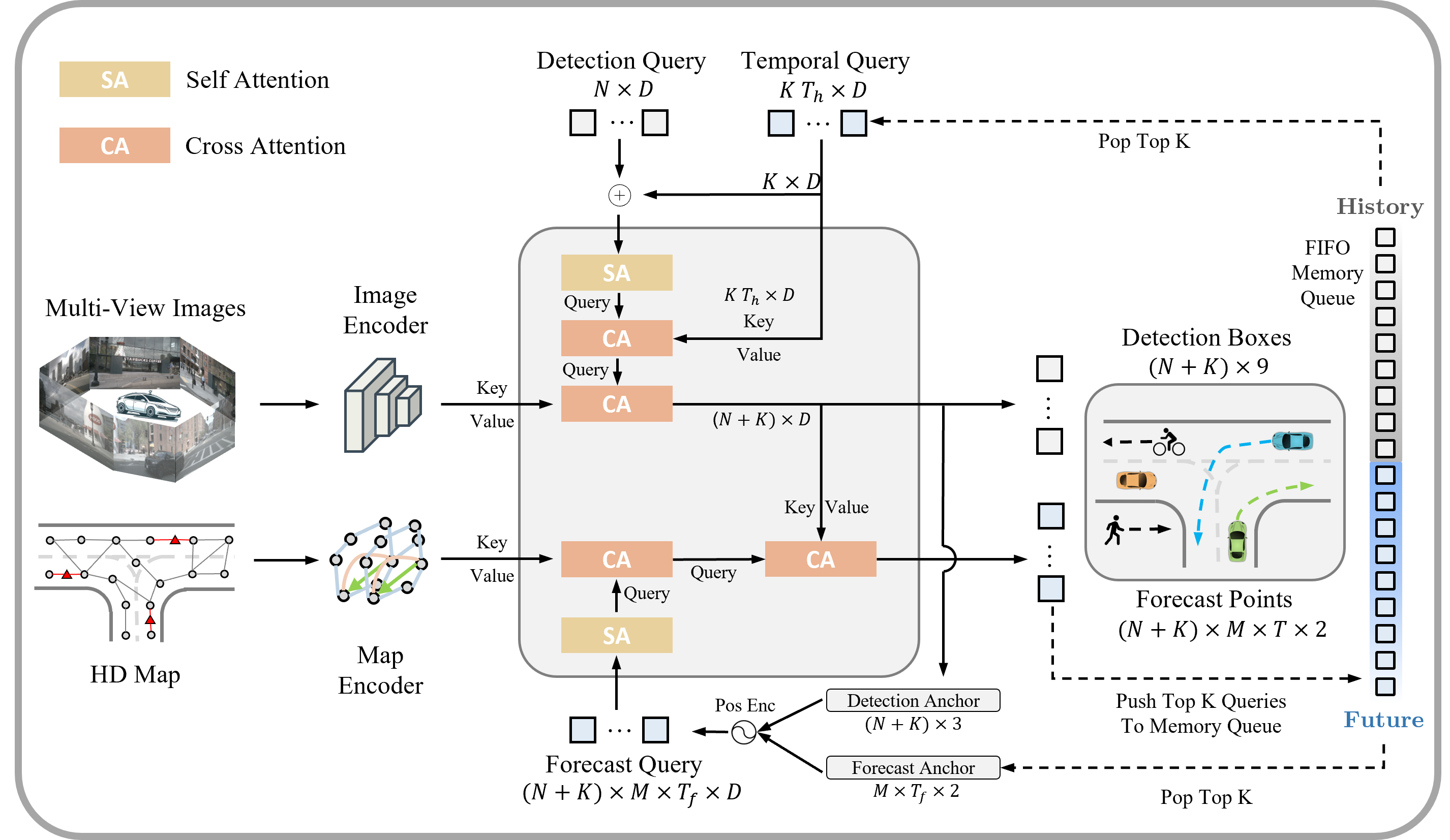} 
  \caption{Overall architecture for the proposed ForeSight method. The model takes multi-view images and an optional HD map as inputs and encodes them separately to produce scene embeddings. Detection queries are combined with a subset of temporal queries from a memory bank and attend to themselves, other temporal queries, and the image embeddings to produce detected objects. The detection positions are combined with forecast anchors to produce forecast queries and attend to themselves, map embeddings, and detection queries to produce future forecasts. Finally, the forecast queries are next pushed to the memory bank for future use.}
  \label{fig:arch}
\end{figure*}

ForeSight extends query-based temporal 3D object detectors~\cite{wang2023exploring} by introducing a forecasting transformer for joint detection and prediction. Streaming query propagation efficiently leverages temporal information, enabling queries to persist within the model over time. ForeSight processes multi-view camera images to detect surrounding objects and generate multi-modal forecasts over future time steps, optionally incorporating high-definition (HD) maps.

An overview of ForeSight is shown in Figure~\ref{fig:arch}. It consists of two primary components: scene encoders for feature extraction and a transformer network for joint streaming detection and forecasting. Section~\ref{sec:embed} discusses the embedding extraction from multi-view images and HD maps. Section~\ref{sec:arch} details the ForeSight transformer architecture for predicting detections and forecasts, and how queries are initialized and propagated. Finally, Section~\ref{sec:learn} describes the optimization approach to jointly train our model end-to-end, including the losses and data augmentation strategy.

\subsection{Scene Embedding Extraction}
\label{sec:embed}

The first stage of ForeSight involves extracting scene embeddings from multi-view images and HD maps. Multi-view images provide the primary sensor input for perception, while HD maps add context for forecasting object motion. The images and HD maps are encoded separately into scene embeddings by individual encoders to be later attended to for detection and forecasting. 

\noindent \textbf{Image Encoder.} At each time step, multi-camera images are processed by an image backbone network (e.g., ResNet-50 \cite{he2016deep}), and extract the visual scene features of each camera view. Following previous works \cite{liu2022petr}, each image feature is fused with a 3D positional embedding to generate 3D position-aware features. These features are then used as a sequence of image feature tokens $\mathbf F_I$ passed to our joint detection and forecasting transformer.

\noindent \textbf{Map Encoder.} If available, HD maps can be encoded to provide road context information, including lanes, traffic signs, and drivable areas. Following the vectorized representation~\cite{gao2020vectornet,liang2020learning}, we represent the HD map as a directed lane graph $\mathcal{G}(V, E)$, where the lanes are broken up into polylines to form the graph nodes $V$, and lane connectivity constructs the graph edges $E$. At each time step, we consider the nearby map elements within an inflated region around the ego agent to account for possible map interaction outside the detection range.
Following \cite{deo2022multimodal}, we first encode the lane polyline sequence according to their connectivity using multi-layer gated recurrent units (GRU), then we apply a graph attention network (GAT) on these lane features to further aggregate local context from the graph structure. We treat lane graph node embedding as a map feature tokens $\mathbf{F}_M$ for later use in our architecture.

\subsection{The ForeSight Transformer}
\label{sec:arch}

Our transformer-based approach centers on two main concepts: (1) a complete process for query initialization, propagation, and exploitation, and (2) an iterative application of self-attention layers to capture intra-query interactions, along with cross-attention layers to aggregate information from scene embeddings. This architecture enables robust integration of temporal information and facilitates interaction and joint learning of detection and forecasting tasks.

\noindent \textbf{Joint Streaming Memory Queue.} A joint memory management and propagation system enables streaming detection and forecasting. We use a joint memory of shape $\mathbf{q}_{mem} \in \mathcal{R}^{K \times T_h \times T_f \times D}$ to represent the top-$K$ objects across $T_h$ historic frames and $T_h$ future frames with a hidden dimension $D$. This queue holds the highest confidence past detection and associated forecast queries, allowing the model to leverage multiple future hypotheses as priors.
At each step, the queue is shifted along the dimension $T_h$ to align with the current time. The corresponding 3D positions, $\mathbf p_q$, of the detection and forecast queries are maintained in the ego reference frame and updated across time using the ego vehicle transformation matrix, $\mathbf T_{ego}$, by
\begin{equation}
    \begin{bmatrix} \mathbf p_{q,t}\\ 1 \end{bmatrix} =  \mathbf T_{ego,t}^{-1} \mathbf T_{ego,t-1} \begin{bmatrix} \mathbf p_{q,t-1}\\ 1 \end{bmatrix}
\end{equation}
We then maintain the query memory in a first-in-first-out manner (FIFO) by appending new decoder outputs and popping the oldest features. While introducing a future queue increases memory requirements, the computation to process it is nearly the same since only $K \times T_h$ out of $K \times T_h \times T_f$ are used for inference at the current frame.

\noindent \textbf{Detection Query Initialization.} 
Following past transformer-based detection methods~\cite{wang2022detr3d,liu2022petr}, we use $N$ detection queries, $\textbf{q}_{det}\in \mathcal{R}^{N\times D}$, to represent and locate objects within the input data. Specifically, at each time step, the detection queries $\textbf{q}_{det}$ are initialized as positional encodings (PE) of a set of 3D object anchor positions $\textbf{x}_{det}\in\mathcal{R}^{N\times 3}$:
\begin{equation}
\label{eq: pos enc}
\mathbf{q}_{det} = \operatorname{MLP}\left(\operatorname{PE}\left(\mathbf{x}_{det}\right)\right)
\end{equation}
where $\mathbf{x}_{det}$ is initialized from a random uniform distribution across the normalized detection range in local 3D space. 

In addition to detection queries, we also leverage temporal memory queries at each timestep, $\mathbf{q}_{temp} \in \mathcal{R}^{K \times T_h \times D}$, from the historical horizon $T_h$. Past objects provide valuable cues about the presence and positions of objects in the current frame. Past methods in both temporal detection\cite{han2024exploring} and tracking \cite{meinhardt2022trackformer} rely on only initializing queries from the top-k last frame outputs. In contrast, our approach takes the top-k from all past queries within $T_f$ frames forecasted to the current time. Since our queries can be initialized over multiple frames in the past, this enables our model to initialize temporal detection queries across occlusions.

\noindent \textbf{Forecast-Aware Detection Transformer.}
The detection queries, $\mathbf{q}_{det}$, and temporal queries, $\mathbf{q}_{temp}$, are then fused to aggregate information from image feature tokens $\mathbf{F}_I$ to detect objects in the current frame.  Specifically, our approach utilizes temporal queries from the past historical frame, $\mathbf{q}_{temp}^{t-1} \in \mathcal{R}^{K \times D}$, and concatenate them with the detection queries, $\mathbf{q}_{det}$. Here, temporal queries carry forward information on objects detected in the most recent past frame, while detection queries are dedicated to identifying newly appearing objects. These concatenated queries, in the shape of $\mathcal{R}^{(N+K)\times D}$, are then fed into self-attention layers to capture intra-query interactions, and into cross-attention layers to attend to all temporal queries in the full historic horizon $\mathbf{q}_{temp}$. Additional cross-attention layers are then applied, where the queries aggregate information from image feature tokens $\mathbf{F}_I$. The resulting queries are refined by successive transformer layers and decoded as bounding boxes $\mathcal{R}^{(N+K)\times 9}$, which include 3D coordinates, box sizes, heading, classes, and confidence.  

\noindent \textbf{Forecast Query Initialization.}
For each decoded detection, ForeSight forecasts its states over the future horizon $T_f$. First, we extract the 2D coordinates and heading from the decoded bounding boxes to form detection anchors of shape $\mathcal{R}^{(N+K) \times 3}$, representing the current states of detected objects. Next, to produce multi-modal forecasts, we follow \cite{shi2022motion, chai2019multipath, hu2023planning} by creating forecast anchors of shape $\mathcal{R}^{M \times T_f \times 2}$. A subset of these anchors are generated via k-means clustering on ground-truth trajectory data to represent trajectory points of distinct modalities. Our work extends this approach further by adding a set of temporal anchors as an additional modality from the previous highest confidence forecast trajectory output. This allows the forecasting model to improve temporal stability and reuse previous computations. Finally, we combine the detection anchors in shape $\mathcal{R}^{(N+K) \times 3}$ with forecast anchors in shape $\mathcal{R}^{M \times T_f \times 2}$, apply positional encodings as in Eq~\ref{eq: pos enc}, and produce initial forecast queries of shape $\mathcal{R}^{(N+K) \times M \times T_f \times D}$.

\noindent \textbf{Joint Streaming Forecast Transformer.}
To perform forecasting, the initial forecast queries first self-attend to themselves, and then cross attend to temporal forecast queries. The queries can then optionally cross-attend to the map feature tokens $\mathbf{F}_{M}$, integrating static scene graph information to enhance trajectory predictions. Specifically, each agent's forecasting queries locally attend to $N_{map}$ elements nearest the agent for efficient agent-map interaction modeling. Lastly, they cross-attend to the processed detection queries, before being decoded into future waypoints of size $\mathcal{R}^{(N+K) \times M \times T_f \times 2}$ and associated confidence after the final layer. By attending to scene embeddings, past temporal detection queries, and past temporal forecast queries, the forecast transformer provides robust trajectory forecasts consistent with the current detections and past forecasts, and accounts for complex spatial patterns. This enables ForeSight to jointly stream historical memory from both detection and forecasting memory queues for query initialization and cross attention. Lastly, the queries are decoded and the top $K$ forecasts are pushed into the temporal query bank for use at future times.

\subsection{Optimization Approach}
\label{sec:learn}

The ForeSight framework is trained using a multi-task objective to optimize both detection and forecasting simultaneously. For detection, a combination of focal loss \cite{lin2017focal} for classification and L1 loss for bounding box regression is used. For forecasting, the model minimizes the trajectory prediction error using an average displacement error (ADE), final displacement error (FDE), and classification loss for the mode closest to the ground truth following \cite{hu2023planning}. In addition, an auxiliary loss is used to supervise ROI feature extraction which also contains a classification and regression loss on targets in the 2D image space \cite{wang2023focal}. The joint loss function is formulated as:
\begin{equation}
    \mathcal{L}_{\mathrm{ForeSight}}=\lambda_{\text {det }} \mathcal{L}_{\text {det }}+\lambda_{\text {forecast }} \mathcal{L}_{\text {forecast}}+\lambda_{\text{aux}} \mathcal{L}_{\text{aux}}
\end{equation}
where $\lambda_{\text{det}}$, $\lambda_{\text{forecast}}$, and $\lambda_{\text{aux}}$ are balancing weights. The forecasting targets are assigned to detections using the nearest ground truth object up to a threshold of 2 meters. This unified loss and assignment encourages the model to leverage shared scene embeddings effectively while maintaining high performance in both detection and forecasting tasks. By adopting this holistic design, the streaming ForeSight architecture ensures tight integration between detection and forecasting, enabling more accurate and consistent scene understanding over time.

\section{Experiments}
\label{sec:experiments}

\begin{table*}[!htb]
\centering
\resizebox{1\linewidth}{!}{
\begin{tabular}{@{}l|c|c|c|c|c|c|c|c|c|c|c@{}}
\toprule
\textbf{Methods}                     & \textbf{Backbone} & \textbf{Image Size} & \textbf{Frames} & \textbf{mAP$\uparrow$} & \textbf{NDS$\uparrow$} & \textbf{mATE$\downarrow$} & \textbf{mASE$\downarrow$} & \textbf{mAOE$\downarrow$} & \textbf{mAVE$\downarrow$} & \textbf{mAAE$\downarrow$} & \textbf{FPS$\uparrow$} \\ \hline
BEVDet4D~\cite{huang2022bevdet4d}    & ResNet50          & $256 \times 704$    & 2               & 0.322                  & 0.457                  & 0.703                     & 0.278                     & 0.495                     & 0.354                     & 0.206                     & 16.7                   \\
PETRv2~\cite{liu2023petrv2}          & ResNet50          & $256 \times 704$    & 2               & 0.349                  & 0.456                  & 0.700                     & 0.275                     & 0.580                     & 0.437                     & 0.187                     & 18.9                   \\
BEVDepth~\cite{li2023bevdepth}       & ResNet50          & $256 \times 704$    & 2               & 0.351                  & 0.475                  & 0.639                     & 0.267                     & 0.479                     & 0.428                     & 0.198                     & 15.7                   \\
BEVStereo~\cite{li2023bevstereo}     & ResNet50          & $256 \times 704$    & 2               & 0.372                  & 0.500                  & 0.598                     & 0.270                     & 0.438                     & 0.367                     & 0.190                     & 12.2         \\
BEVPoolv2~\cite{huang2022bevpoolv2}  & ResNet50          & $256 \times 704$    & 9               & 0.406                  & 0.526                  & 0.572                     & 0.275                     & 0.463                     & 0.275                     & \underline{0.188}                     & 16.6                   \\
VideoBEV~\cite{han2024exploring}     & ResNet 50         & $256 \times 704$    & 2               & 0.422                  & 0.535                  & \textbf{0.564}            & 0.276                     & \underline{0.440}            & 0.286                     & 0.189                     & -                      \\
SOLOFusion~\cite{park2022time}       & ResNet50          & $256 \times 704$    & 17              & 0.427                  & 0.534                  & \underline{0.567}           & 0.274                     & 0.511                     & \textbf{0.252}            & \textbf{0.181}           & 11.4          \\
StreamPETR*~\cite{wang2023exploring} & ResNet50          & $256 \times 704$    & 4               & \underline{0.445}        & \underline{0.538}                  & 0.640                     & \underline{0.267}           & 0.443                     & 0.259           & 0.206                     & \textbf{31.7}                   \\
ForeSight (ours)                     & ResNet50          & $256 \times 704$    & 4               & \textbf{0.466}         & \textbf{0.560}        & 0.614                     & \textbf{0.266}            & \textbf{0.370}           & \underline{0.258}                     & 0.201                     & \underline{23.5}                   \\ \midrule
BEVDepth~\cite{li2023bevdepth}                             & ResNet101         & $512 \times 1408$   & 2               & 0.412                  & 0.535                  & \underline{0.565}                     & 0.266                     & 0.358                     & 0.331                     & \underline{0.190}                     & -                        \\
BEVFormer~\cite{li2022bevformer}                             & ResNet101-DCN         & $900 \times 1600$   & 4               & 0.416                  & 0.517                  & 0.673                     & 0.274                     & 0.372                     & 0.394                     & 0.198                     & 3.0                        \\
Sparse4D~\cite{lin2023sparse4d}                           & ResNet101-DCN         & $900 \times 1600$   & 4               &  0.436                   & 0.541                  & 0.633                     & 0.279                     & 0.363                     & 0.317                     & \textbf{0.177}                     & 4.3                    \\
SOLOFusion~\cite{park2022time}                           & ResNet101         & $512 \times 1408$   & 17              & 0.483                  & \underline{0.582}                  & \textbf{0.503 }                    & \underline{0.264}                     & 0.381                     & \underline{0.246}                     & 0.207                     & -                        \\
StreamPETR*~\cite{wang2023exploring}                          & ResNet101         & $512 \times 1408$   & 4               & \underline{0.486}                       &        0.578                &       0.600                    &         0.270                  &         \textbf{0.343}                  &            0.248               &             0.192              & \textbf{6.4}                        \\
ForeSight (ours)                            & ResNet101         & $512 \times 1408$   & 4               & \textbf{0.502}                       &         \textbf{0.589}               &        0.578                   &           \textbf{0.255}                &          \underline{0.346}                 &           \textbf{0.238}                &           0.193                &  \underline{5.1}                      \\ \midrule
StreamPETR*~\cite{wang2023exploring}      & V2-99          & $900 \times 1600$                         & 4               & 0.479         &       0.569        & 0.622          & 0.263          & \textbf{0.364}          & 0.256          &           0.194 & - \\
ForeSight (ours)             & V2-99          & $900 \times 1600$                         & 4               & \textbf{0.489}         &       \textbf{0.577}        & \textbf{0.598}          & \textbf{0.259}          & 0.376          & \textbf{0.250}          & 0.194  & -   \\ \bottomrule
\end{tabular}
}
\caption{Detection performance on NuScenes validation set, against comparable SOTA vision-based methods. ForeSight delivers competitive performance compared to SOTA methods, achieving the highest mAP. Notably, ForeSight surpasses the second-best detection baseline, StreamPETR, with a notable 2.1\% improvement in mAP. *Results generated with the same number of queries and frames as ForeSight for a fair comparison.}
\label{tab:val}
\end{table*}

\subsection{Experiment Setup}
\textbf{Datasets.} We evaluated ForeSight on the NuScenes 3D detection dataset \cite{caesar2020nuscenes}. NuScenes consists of 1,000 scenarios; each scenario is roughly 20s long, with a sensor frequency of 20 Hz. Each sample contains images from 6 cameras at different viewing angles around the vehicle. Camera parameters including intrinsics and extrinsics are available. NuScenes provides detection annotations every 0.5s; in total there are 28k, 6k, and 6k annotated samples for training, validation, and testing, respectively. There are 10 total classes to evaluate against, including multiple vehicle, pedestrian, and cyclist classes. 

\noindent The official NuScenes forecasting benchmark only provides forecast ground truth for hand selected and highly-curated objects, while we aim at detecting and forecasting all objects in the sensor data. Therefore, following past works \cite{pang2023standing, hu2023planning, gu2023vip3d} we construct a NuScenes forecasting dataset for training and evaluation from the labeled sensor data. We leverage the available unique track identities for each detection to construct ground truth future forecasts for every detection box in the dataset. We follow the standard practice of using 12 future frames that spans 6 seconds for our forecasting data.

\noindent\textbf{Metrics.} In our experiments, we focus on key detection metrics such as mean Average Precision (mAP) to assess detection capabilities. Following the official nuScenes evaluation protocol with a 2-meter true positive distance threshold, we also provide mean Average Translation Error (mATE), mean Average Scale Error (mASE), mean Average Orientation Error (AOE), mean Average Velocity Error (mAVE), and mean Average Attribute Error (mAAE). To capture all aspects of the detection task, we report the NuScenes Detection Score (NDS), a consolidated scalar metric. We include the Frames Per Second (FPS) to evaluate the computation efficiency.

\noindent Forecasting metrics include minimum Average Displacement Error (minADE), minimum Final Displacement Error (minFDE), Miss Rate (MR), and End-to-End Prediction Accuracy (EPA)\cite{gu2023vip3d}. We follow common motion forecasting benchmarks \cite{chang2019argoverse,wilson2023argoverse,sun2020scalability} to compute metrics over the top 6 different trajectory mode hypotheses by only evaluating the minimum distance trajectory from the ground truth.  The maximum distance for the miss rate is taken as 2 meters. To compute the forecasting metrics, we first associate the current detections to the closest ground truth bounding boxes with a maximum distance threshold of 1 meter.

\noindent\textbf{Implementation Details.} We conduct experiments with the popular image backbones, ResNet50, ResNet101 \cite{he2016deep}, V2-99 \cite{lee2019energy}, ViT-L~\cite{dosovitskiy2020image} to facilitate comparisons to other methods. Following past works \cite{wang2023exploring, li2022bevformer, liu2022petr} the ResNet model is initialized from pretrained weights trained on NuImages\cite{caesar2020nuscenes}, V2-99 on DD3D \cite{park2021pseudo}, and ViT-L from EVA-02~\cite{fang2024eva} . The map encoder was adapted from PGP \cite{deo2022multimodal} and converted from an agent-level encoder to a scene-level encoder. We utilize $N=300$ detection queries and $K=128$ temporal queries for a total of $428$ queries that are decoded into detection bounding boxes. In the first frame, temporal queries are replaced with additional detection queries, acting as a single-frame detector. We use a temporal history of $T_H=4$ frames and future forecast horizon for $T_f=12$ frames. The forecast queries are generated with $M=6$ forecast modes and permuted with the 428 detection queries across 12 future frames for a total of 30,816 forecast queries that are decoded into multimodal forecast waypoint positions. We also train using detection and forecast queries denoising \cite{wang2023focal, wang2023exploring} during training to enrich supervision.

We use 6 transformer layers to decode detection boxes and 3 transformer layers for the forecast positions. The dimension $D$ of all embeddings is 256. For efficiency, only the $N_{map}=16$ map nodes closest to each detection are used for cross-attention with the map embeddings. For the weights in our multi-task objective, we use $\lambda_{det} = 1, \lambda_{forecast} = 2$. We train our model for 20 epochs using the AdamW \cite{loshchilov2017decoupled} optimizer with a batch size of 16 and base learning rate of $4 \times 10^{-4}$. A cosine annealing learning rate decay is used starting at $10^{-3}$ with $500$ warm-up iterations and a weight decay of $10^{-2}$ is applied. 

\subsection{Comparison Against State-of-the-Art}
We evaluate our detection and forecasting performance on NuScenes against comparable state-of-the-art (SOTA) vision-based methods, which are shown in Table~\ref{tab:val} and Table~\ref{tab:val_forecast}, respectively.

\noindent\textbf{Detection Performance.} ForeSight outperforms state-of-the-art (SOTA) multi-view 3D detection methods, achieving the highest mAP and NDS. Notably, ForeSight surpasses the best detection baseline, StreamPETR, with a 2.1\% improvement in mAP. This achievement highlights ForeSight's enhanced precision and robustness in object detection and temporal reasoning. These findings underscore the importance of leveraging feedback from forecasting to improve detection capabilities.

\begin{table}[!htb]
\centering
\resizebox{1\linewidth}{!}{
\begin{tabular}{@{}l|c|c|c|c@{}}
\toprule
\textbf{Method}         & \textbf{Base}        & \textbf{EPA} $\uparrow$ & \textbf{mAP} $\uparrow$ & \textbf{minADE} $\downarrow$ \\ \midrule
ViP3D~[11]   & R50   & 0.226  & 0.284     & 2.05  \\ 
StreamPETRcv~[57]*  & R50  & 0.387  & \underline{0.445} & \underline{1.75} \\ 
ForeSight (ours)  & R50  & \textbf{0.499} & \textbf{0.466} & \textbf{0.709} \\ \midrule
VAD~[21]\dag      & R50 & 0.516      & 0.354 &  0.511   \\
ForeSight (ours) \dag      &  R50      & \textbf{0.606}  & \textbf{0.713} & \textbf{0.379}    \\ \midrule 
PnPNet~[16, 31] & R101 & 0.222  & - & 1.15   \\
StreamPETRcv~[57]*      & R101             &   0.453  & \underline{0.486} & 1.64   \\
UniAD~[16] & R101  & \underline{0.456}  & 0.382 & \underline{0.708}   \\
ForeSight   (ours)     & R101       & \textbf{0.549}  & \textbf{0.502} & \textbf{0.689} \\ \midrule 
StreamPETRcv~[57]*  & V2-99  & 0.441 & 0.479 &   1.71         \\ 
ForeSight (ours)  & V2-99  & \textbf{0.535} & \textbf{0.489} &    \textbf{0.693}        \\ \midrule
PF-Track~[39] \ddag & V2-99 & 0.452      & 0.327 & 1.77   \\ 
ForeSight (ours) \ddag &  V2-99 & \textbf{0.574}  & \textbf{0.498} & \textbf{0.578}   \\ \midrule 
StreamPETRcv~[57]*  & ViT-L  & 0.486 & 0.521 &  1.38             \\ 
ForeSight (ours)  & ViT-L  & \textbf{0.545} & \textbf{0.543} &     \textbf{0.687}         \\ \bottomrule
\end{tabular}
}
\caption{End-to-end detection and forecasting performance on NuScenes val set. ForeSight achieves highest detection and forecasting performance against comparable methods. \dag 3s forecast horizon, 15x30m perception range, and 4 classes. \ddag 4s forecast horizon and 8 classes. *Detector with constant velocity forecast.}
\label{tab:val_forecast}
\end{table}

\noindent\textbf{Forecast Performance.}
We compare our forecasting results to other open-sourced forecasting methods that operate directly from perception inputs. For end-to-end prediction accuracy~\cite{gu2023vip3d} (EPA) our framework achieves the best performance, beating the SOTA end-to-end method UniAD by a substantial 9.3\% due to higher forecast hit rate and lower false positive detection rate. This highlights the broad applicability of ForeSight in end-to-end autonomous driving systems by joint streaming detection and forecasting for superior performance. 

Foresight improves on minADE prediction performance across a much larger set of real trajectories, enabled by more accurate perception matching more detections to the ground truth. This underscores a key limitation of directly comparing forecasting performance across different perception approaches. Stronger detectors capture more challenging objects leading to a a more diverse and difficult prediction sample set, potentially skewing prediction comparisons. Another challenge is the limited number of forecasting-from-sensor methods, each relying on different image backbones and perception networks, which can influence downstream forecasting performance. We adapt our evaluation to match prior works with varying configurations (e.g., backbones, perception range, forecast horizon, classes) in order to establish a fair and unified benchmark remains an important direction for future research.

\begin{figure*}[!ht]
  \centering
  \includegraphics[width=\columnwidth]{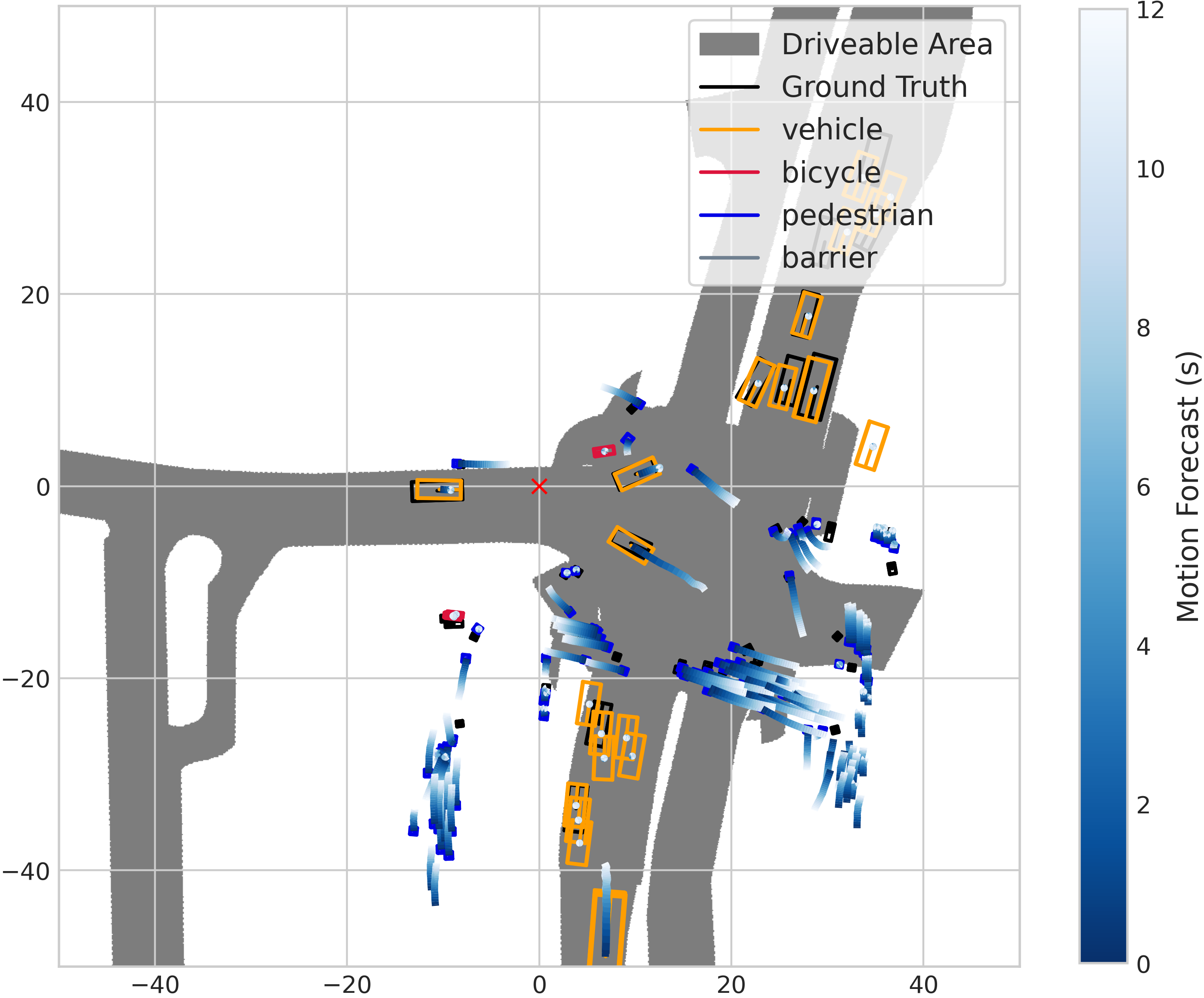} 
\includegraphics[width=\columnwidth]{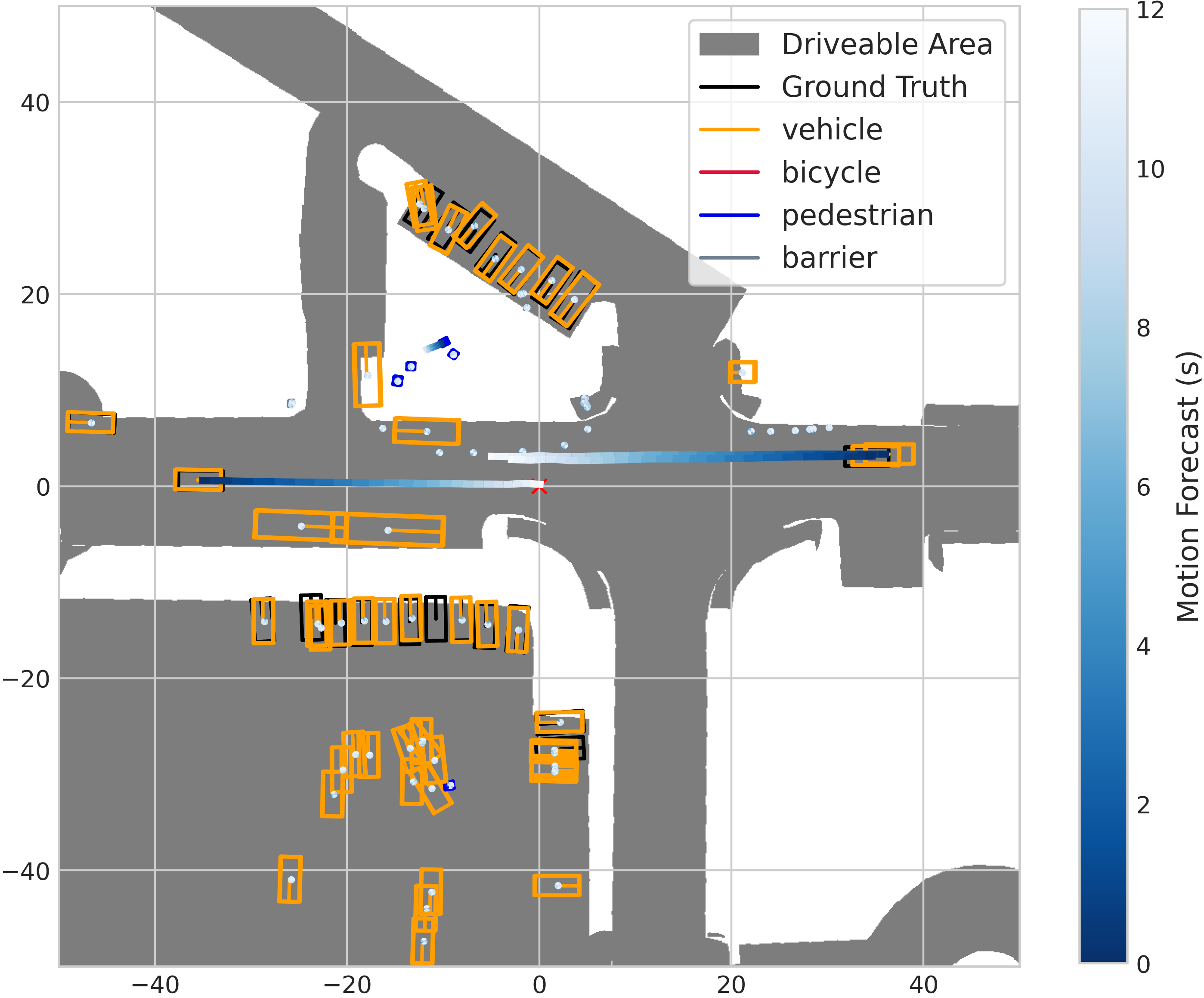} 
  \caption{Qualitative case with many pedestrians crossing (left) and occluded parked cars (right). ForeSight detects the crossing pedestrians and forecasts their motion while predicting that the vehicles will stay stationary to yield to the pedestrians. ForeSight can also detect limited visibility vehicles using previous observations.}
  \label{fig:qual}
\end{figure*}

\subsection{Ablation Study and Analysis}

We investigate the design choices made in ForeSight's architecture and present the analysis in Table~\ref{tab:val_abl1}. We start with a baseline detector based on the PETR model \cite{liu2022petr} with the ResNet-50 backbone and subsequently add the components that make up our approach of bidirectional learning for joint detection and forecasting.

\begin{table}[!htb]
\centering
\resizebox{1\linewidth}{!}{
\begin{tabular}{@{}c|c|c|c|c|c|c|c@{}}
\toprule
\multirow{2}{*}{\begin{tabular}[c]{@{}c@{}}\textbf{Detect}\\ \textbf{Frames}\end{tabular}} & \multicolumn{2}{|c|}{\textbf{Forecast}} & \multicolumn{2}{|c|}{\textbf{Propogate}} & \multirow{2}{*}{\begin{tabular}[c]{@{}c@{}}\textbf{HD}\\ \textbf{Map}\end{tabular}} & \multicolumn{2}{c}{\textbf{Metrics}} \\ \cmidrule(l){2-5}  \cmidrule(l){7-8}
 & \multicolumn{1}{c|}{\textbf{TF}} & \multicolumn{1}{c|}{\textbf{DF}} & \multicolumn{1}{c|}{\textbf{DP}} & \multicolumn{1}{c|}{\textbf{FP}} & & \multicolumn{1}{c|}{\textbf{mAP} $\uparrow$} & \textbf{minADE} $\downarrow$ \\ \midrule
1 & &  &  &  & & 0.361 & - \\
1 & \checkmark &    &    & & &   0.361   &   2.83    \\
1 &   & \checkmark  &     & & &  0.361   &  1.05   \\ 
4 &   &     &  \checkmark    & & &   0.440     &   -  \\ 
4 &   & \checkmark &   & \checkmark & & \underline{0.463} & \underline{0.735} \\
4 &   & \checkmark &   & \checkmark & \checkmark & \textbf{0.466} & \textbf{0.709} \\ \bottomrule
\end{tabular}
}
\caption{Ablation of design decisions on nuScenes validation set. Starting from a baseline single-frame detector \cite{liu2022petr} we compare tracking-based forecasting (TF) vs detection-based forecasting (DF), direct propagation (DP) vs forecast-based propagation (FP), and offline HD map usage with the full ForeSight system in the last row.}
\label{tab:val_abl1}
\end{table}

\noindent\textbf{Tracking-Free vs. Tracking-Aware Forecast.}
Tracking-based forecasting is a common forecasting approach \cite{deo2022multimodal} which is trained on ground truth trajectory histories and therefore requires accurate track data to produce good results. When applying these models to real tracks generated from our detections, we find that performance is significantly degraded due to the lack of robustness to errors in the tracks. Instead, we rely on a detection-based forecasting (DF) approach which substantially improves minADE by 1.78 m compared to a similar tracking-based forecasting (TF) approach trained on the nuScenes prediction dataset. By removing the tracking bottleneck, the forecasting model is significantly more robust to perception errors and shows better results for the challenging driving scenarios contained in the forecasting data we generated from NuScenes. ForeSight’s tracking-free design facilitates a more streamlined information flow within the perception-forecasting pipeline.

\noindent\textbf{Impact of HD Map.} ForeSight is optionally able to make use of an HD map for forecasting via cross-attention to encoded map queries. While detection does not directly leverage HD maps, it benefits indirectly through improved forecast propagation. Table~\ref{tab:val_abl1} shows the HD map usage can slightly improve forecasting performance with a reduction of 0.026 minADE, highlighting its role in enhancing forecasting and indirectly improving detection.

\noindent\textbf{Impact of Bidirectional Propagation.} We evaluate the effectiveness of ForeSight’s bidirectional query propagation, which enhances temporal reasoning and maintains query association across frames. By propagating queries between detection and forecasting, ForeSight outperforms traditional sparse-query models that propagate only forward. Specifically, using naïve direct detection query propagation, following the StreamPETR \cite{wang2023exploring} approach, we achieve a detection mAP of 0.440, a 2.6\% mAP improvement over the baseline single-frame baseline detector. We show that we can further improve performance on the detection task to 0.463 mAP by propagating the forecast query back to the detection task. Together these enhancements show a clear benefit to ForeSight's bidirectional query propagation and joint learning for detection and forecasting. 

\subsection{Qualitative Results}

Qualitative results are provided to visualize the outputs of ForeSight and better understand the model's ability to process temporal information and use it effectively. Figure~\ref{fig:qual} shows a case where many static vehicles are parked. By using past detections and correctly predicting that they will be stationary the model can accurately detect parked cars even when they are heavily occluded. In addition, the model correctly forecasts moving vehicles and stationary vehicles. Figure~\ref{fig:qual} also shows complex behavior with a large group of pedestrians. We can see that the model can accurately forecast the social interaction between these clusters of dynamic objects and correctly predicts their motion together. Even though many of these objects are obstructed at the given time, they have been seen in the past and the model correctly detects they are still present in the group. In addition, the correct conditional motion is predicted such that the vehicles in the scene are yielding to the pedestrians and cyclists. These visualizations further demonstrate ForeSight's strength in detecting partially obstructed objects and predicting object trajectory, reinforcing its innovative ability to unlock improved performance. 
\section{Conclusion}
\label{sec:conclusion}

In this work, we present ForeSight, a unified framework for joint detection and forecasting in vision-centric autonomous driving. Our approach bridges the gap between perception and prediction by introducing a bidirectional learning mechanism and joint streaming memory, where object queries propagate seamlessly between detection and forecasting. This design not only enhances detection accuracy in complex driving scenarios but also improves forecasting robustness to real-world perception uncertainties. Extensive evaluations on the nuScenes dataset validate ForeSight’s effectiveness, achieving state-of-the-art performance for detection. With bidirectional query propagation and a tracking-free architecture, ForeSight advances autonomous vehicle perception, enhancing efficiency, reliability, and safety in real-world navigation. 
{
    \small
    \bibliographystyle{ieeenat_fullname}
    \bibliography{main}
}
\clearpage
\appendix 
\twocolumn[
\begin{@twocolumnfalse}
\centering
\section*{Supplementary Material}
\vspace{0.5em}
\end{@twocolumnfalse}
]

The supplementary material includes four key sections designed to provide a comprehensive evaluation of ForeSight. Section A presents additional experimental analysis on performance in challenging scenarios.Section B provides additional ablation results for design decisions. Section C offers a detailed class-wise analysis, highlighting ForeSight's superior performance across diverse object categories, including challenging cases such as trailers and motorcycles. Section D discusses the method's limitations, including dependencies on high-quality data, challenges in adverse weather, the need for standardized forecasting benchmarks, and adaptability to dynamic scenes, while proposing avenues for future improvement.



\section{Detection Performance Analysis}

We highlight ForeSight’s enhanced performance on challenging scenarios like low-visibility and occluded objects, critical for safe autonomy. While these cases are a small portion of the data, they are crucial long-tail challenges. Leveraging past forecasts, our model can more effectively detect objects with limited sensor visibility and coverage. Quantitative analysis with the ResNet-50 configuration of ForeSight confirms improved performance on low-visibility objects, with $<$40\% of the object visible, and achieve a 0.9\% higher Average Recall (AR) on these challenging objects relative to the baseline as seen in Table~\ref{tab:vis_ablation}.


\begin{table}[!htb]
\resizebox{0.95\linewidth}{!}{
\begin{tabular}{l|c|c|c|c}
\hline
\textbf{Methods} & \textbf{0-0.4} & \textbf{0.4-0.6} & \textbf{0.6-0.8} & \textbf{0.8-1.0}\\ \hline
StreamPETR~[59] &  0.401 & 0.455 & 0.448 & 0.468  \\
ForeSight (ours) & \textbf{0.410} & \textbf{0.461} & \textbf{0.451} & 0.468 \\ \hline
\end{tabular}
}
\caption{Detection performance using average recall across different object visibility levels.}
\label{tab:vis_ablation}
\end{table}

\section{Additional Ablations}

\textbf{Ablation for historical frame count.}
Table~\ref{tab:val_abl2} presents additional experiments varying historical frames. Performance improves with more frames, with diminishing gains beyond 4 frames. In addition, as many prior works also use 4 frames, this choice enables fair comparison.

\begin{table}[!htb]
\centering
\resizebox{0.8\linewidth}{!}{
\begin{tabular}{@{}c|c|c|c@{}}
\toprule
\textbf{Frames} & \textbf{Det. Queries} & \textbf{CA Queries} & \textbf{mAP} $\uparrow$  \\ \midrule
0 & 900 (900+0) & 0 & 0.361   \\ 
1 & 900 (644+256) & 256    &   0.419  \\ 
2 & 900 (644+256) & 512    &  0.448     \\ 
4 & 900 (644+256) & 1024 & 0.466 \\ 
8 & 900 (644+256) & 2048 &  \textbf{0.467}    \\ \bottomrule
\end{tabular}
}
\caption{Ablation of historic frames, detection queries (initialized + propagated), and cross attention (CA) historic queries.}
\label{tab:val_abl2}
\end{table}

\textbf{Ablation replacing historical queries with spatial queries.}
In Table~\ref{tab:val_abl3}, replacing historical queries with spatial ones (first row) while keeping 900 total detection queries results in a reduction of 10.5\% mAP and increase of 3.4 m minADE compared to our model (last row), isolating the benefit of temporal information from query count.

\textbf{Study of the relation between forecasting errors and detection accuracy.}
We conduct additional experiments varying the query propagation mechanism (Table~\ref{tab:val_abl3}). Starting from a detection memory queue with stationary past queries, adding forecasting decoder (FD) layers improves forecasting accuracy, which in turn enhances detection, highlighting the synergy between tasks.

\begin{table}[!htb]
\centering
\resizebox{0.7\linewidth}{!}{
\begin{tabular}{@{}c|c|c@{}}
\toprule
\textbf{Query Propogation} & \textbf{mAP} $\uparrow$ & \textbf{minADE} $\downarrow$ \\ \midrule
Constant Pos. & 0.445 & 4.13  \\
FD 1 Layer    & 0.458 & 0.979  \\
FD 3 Layers   & \textbf{0.466} & \textbf{0.709}\\ \bottomrule 
\end{tabular}
}
\caption{Ablation of query propagation approaches used for historical detection queries.}
\label{tab:val_abl3}
\end{table}

\section{Class-wise Analysis}

In Table~\ref{tab:class}, we present a detailed comparison of class-wise performance for the V2-99 backbone on the NuScenes validation set. We show results for AP at the 2.0m threshold. ForeSight consistently outperforms the baseline comparison across most object classes, demonstrating its robustness and versatility in detecting a diverse range of objects in dynamic driving scenarios.

\begin{table*}[t]
\centering
\resizebox{0.9\linewidth}{!}{
\begin{tabular}{@{}l|c|c|c|c|c|c|c|c@{}}
\toprule
\textbf{Methods} & \textbf{Backbone} & \textbf{Car$\uparrow$} & \textbf{Pedestrian$\uparrow$} & \textbf{Bicycle$\uparrow$} & \textbf{Bus$\uparrow$} & \textbf{Motorcycle$\uparrow$} & \textbf{Trailer$\uparrow$} & \textbf{Truck$\uparrow$} \\ \midrule
StreamPETR*~\cite{wang2023exploring} & V2-99  & 0.810 & 0.729 & 0.603 & 0.710 & 0.627 & 0.366 & \textbf{0.628}  \\
ForeSight (ours)                     & V2-99 & \textbf{0.812} & \textbf{0.731} & \textbf{0.608} & \textbf{0.750} & \textbf{0.638} & \textbf{0.421} & 0.607 \\ \bottomrule
\end{tabular}
}
\caption{Detection performance on NuScenes validation set comparing class-wise performance based on AP with a 2.0 meter distance threshold. We observe that our model improves on nearly all classes of objects and performs significantly better on challenging classes with lower performance such as trailers, motorcycles, and bicycles.}
\label{tab:class}
\end{table*}

Significant improvements are observed in challenging classes such as trailers and motorcycles, where the mAP gains are 15.0\% and 1.8\%, respectively. These categories often suffer from lower detection rates due to less frequent appearance in the data, irregular shapes, and variability in motion patterns. By leveraging enhanced temporal modeling and spatial context, ForeSight can provide a more accurate and reliable detection for these difficult cases.

Additionally, for high-performance classes such as cars and pedestrians, ForeSight continues to achieve competitive or superior results, maintaining its overall edge in performance. This demonstrates that the proposed method not only excels in challenging scenarios but also scales effectively across well-represented object categories.

These results underscore ForeSight's ability to generalize across object types and validate its superiority in real-world applications requiring precise multi-class detection. The improved performance on underrepresented or challenging object classes further highlights the method's enhanced temporal reasoning and adaptability.



\section{Limitations}

\textbf{Limited Comparisons for Forecasting}. Due to the lack of standardized forecasting-from-perception benchmarks, our evaluation relies on adapting the NuScenes detection and tracking dataset following the approach of past works. Direct forecasting comparison to other end-to-end methods is challenging due to differing configurations (e.g. backbones, perception models) and a lack of standardized evaluation frameworks. Future work will explore establishing a common benchmark to evaluate similar methods with common upstream models to provide a proper fair comparison.

\textbf{High-Quality Data Dependancy}. The effectiveness of ForeSight may depend on the quality of the input data. Since we rely on tight coupling of temporal information errors in camera calibration, localization, or map inaccuracies can propagate through the pipeline, potentially degrading both detection and forecasting outcomes. Due to the feedback loop in the pipeline, it could also be more susceptible to these errors that deteriorate performance. The robustness of errors could be explored in future work along with mitigation strategies.

\textbf{Adverse Weather}. A potential limitation of ForeSight could also be sensitivity to adverse weather conditions such as heavy rain, snow, or fog. These conditions can degrade the quality of camera sensor inputs by obscuring object boundaries, reducing contrast, and introducing noise. As ForeSight relies heavily on vision-based perception, any reduction in image quality directly impacts both detection and forecasting accuracy. Future work could explore the integration of additional sensor modalities, such as LiDAR or radar, which are more robust to weather-induced impairments. Robust data augmentation strategies simulating adverse weather conditions during training may also improve the resilience of the model in such challenging environments.

\textbf{Simplified Scene Representations}: The method's reliance on predefined object categories limits its adaptability to environments with novel object classes not represented in training data. The ability to segment or detect map elements online has also been explored in other works and could be integrated into this method instead of using an offline map or no map.

\end{document}